\documentclass[10pt, conference, compsocconf]{IEEEtran}
\usepackage[utf8]{inputenc}
\usepackage{fourier} 

\usepackage{array}
\usepackage{makecell}

\usepackage{amsmath}

\usepackage{balance}  
\usepackage{graphics} 
\usepackage{times}    
\usepackage{url}      
 
\usepackage{graphicx}
\usepackage{caption}
\usepackage[labelformat=simple]{subcaption}
\usepackage{bbold}
\usepackage{multirow}

\usepackage{comment}
\usepackage[ruled]{algorithm2e} 

\usepackage[usenames, dvipsnames]{color}
\SetAlFnt{\small}
\SetAlCapFnt{\small}
\SetAlCapNameFnt{\small}
\SetAlCapHSkip{0pt}
\IncMargin{-\parindent}

\usepackage{tikz}
\usetikzlibrary{fit,shapes.geometric}
\newcounter{nodemarkers}
\newcommand\circletext[1]{%
    \tikz[overlay,remember picture] 
        \node (marker-\arabic{nodemarkers}-a) at (0,1.5ex) {};%
    #1%
    \tikz[overlay,remember picture]
        \node (marker-\arabic{nodemarkers}-b) at (0,0){};%
    \tikz[overlay,remember picture,inner sep=2pt]
        \node[draw,ellipse,fit=(marker-\arabic{nodemarkers}-a.center) (marker-\arabic{nodemarkers}-b.center)] {};%
    \stepcounter{nodemarkers}%
}

\ifCLASSINFOpdf
\else
\fi
\begin{document}
%
\title{Semi-Supervised Convolutional Neural Networks for Human Activity Recognition}


\author{
\IEEEauthorblockN{Ming Zeng, Tong Yu, Xiao Wang, Le T. Nguyen, Ole J. Mengshoel, Ian Lane}
\IEEEauthorblockA{Carnegie Mellon University, Moffett Field, CA 94043}
{\{ming.zeng, tong.yu, xiao.wang, le.nguyen, ole.mengshoel, ian.lane\}}@sv.cmu.edu

}


%


\maketitle

\begin{abstract}
Labeled data used for training activity recognition classifiers are usually limited in terms of size and diversity. Thus, the learned model may not generalize well when used in real-world use cases. 
Semi-supervised learning augments labeled examples with unlabeled examples, often resulting in improved performance.
However, the semi-supervised methods studied in the activity recognition literatures assume that feature engineering is already done. In this paper, we lift this assumption and present two semi-supervised methods based on convolutional neural networks (CNNs) to learn discriminative hidden features. Our semi-supervised CNNs learn from both labeled and unlabeled data while also performing feature learning on raw sensor data. In experiments on three real world datasets, we show that our CNNs outperform supervised methods and traditional semi-supervised learning methods by up to $18\%$ in mean F1-score ($F_m$).
\end{abstract}

\begin{IEEEkeywords}
Human Activity Recognition; Deep Neural Networks; Semi-Supervised Learning; Convolutional Neural Networks
\end{IEEEkeywords}

%
\IEEEpeerreviewmaketitle

\section{Introduction}

Human activity recognition (HAR) is an important application area for mobile, on-body, and worn mobile technologies. Supervised learning for human activity recognition has shown great promise. Among supervised methods, deep neural networks (DNNs) have emerged as a method with much potential, in that they are less dependent on clever feature engineering and has strong generalization ability~\cite{zhang2016understanding} compared to other supervised methods~\cite{lane2015can, ordonez2016deep}. 

Unfortunately, the problem of data labeling remains. Compared to many other machine learning applications, the problem of data labeling for HAR is substantial, since human activity data sets typically (i) have few labeled samples and (ii) are highly personal and varying. 

(i) Activity data sets typically have very few labeled examples for some activities.  Thus, they may not characterize well the distribution of test data collected in different situations than the training data. For example, the labeled training data may only cover walking at certain speeds. In reality, humans walk at a range of speeds. They can walk slowly when being relaxed and can walk very fast when in a hurry. The problem of limited labels is even more severe for models with high parameter complexity, such as deep neural networks.


(ii) Activity data sets are highly personal and varying, because people may perform the same activity in very different ways. For example, what one person considers jogging may be very similar to what another person considers fast walking.  With a model trained only on data where a human walks at normal speed, it is very difficult to correctly predict the behavior of a human walking in a hurry. Walking in a hurry can easily be confused with running, especially when little data of walking in a hurry is collected for training.


To address challenges (i) and (ii), many semi-supervised learning methods have been proposed to leverage the abundance of unlabeled data and provide higher generalizability. Although the labeled data of walking in a hurry may be limited, there are large amounts of \emph{unlabeled} data recording the behavior of walking in a hurry. 
Semi-supervised learning from both labeled and unlabeled data can thus potentially provide better predictions for human walking in a hurry, compared to supervised learning using only labeled data. 

When labeled data is limited, we may potentially improve HAR performance via adjustments to labeled data's feature representations with unlabeled data, so-called feature learning. In contrast, previous semi-supervised HAR approaches usually rely on handcrafted features~\cite{stikic2008exploring, stikic2009multi,yao2016learning}. With hand-crafted features, the benefit of the unlabeled data is limited, since there is no opportunity for feature learning with the unlabeled data. 


In this paper, we study how to train accurate and generalizable DNNs with limited labeled data and large scale unlabeled data for HAR. 
Specifically, we present two semi-supervised deep convolutional neural network methods, the convolutional encoder-decoder (CNN-Encoder-Decoder) and the convolutional ladder network (CNN-Ladder). The contributions of our work are the following. 

\begin{itemize}
\item To our best knowledge, this is the first paper to leverage unlabeled data in CNNs in HAR applications. We utilize unlabeled data in both feature learning and model learning using CNN-Encoder-Decoder and CNN-Ladder architectures for semi-supervised HAR.
\item The presented methods can achieve up to $18\%$ F1-score improvement compared to baseline methods, on three real-world activity recognition datasets.
\item To understand why our methods improve F1-score, we show the importance of adjusting low level features based on unlabeled data in semi-supervised HAR. Besides, we visualize the features in the last layers of CNN-Ladder and CNN to demonstrate that better  high-level features can be learned with unlabeled data added.
\end{itemize}



\section{Related Works}
\label{sec:relatedwork}
In this section, we discuss related work on (i) machine learning in HAR and (ii) semi-supervised learning in HAR.
\subsection{Machine Learning for Activity Recognition}

In early studies of HAR~\cite{bao2004activity}, machine learning models using handcrafted features shows good performance. Raw sensor data is collected from various sensors 
on mobile devices.
From this collected data, handcrafted features are designed using domain knowledge. With the handcrafted features, machine learning models, such as random forest, naive Bayes, or SVMs, are trained and used in HAR.

Designing handcrafted features requires domain knowledge~\cite{bulling2014tutorial}. Therefore, it is desirable to develop a systematic feature learning approach to model the time series signals in HAR~\cite{yang2015deep}. Deep neural networks (DNNs) are emerging feature extraction approaches to HAR, and they have made great advances in many domains~\cite{lecun2015deep}. They are also applied to HAR ({\em e.g.}, ~\cite{plotz2011feature,zeng2014convolutional,hammerla2016deep}). The first HAR deep learning approach~\cite{plotz2011feature} explores unsupervised feature extraction. It outperforms principal component analysis (PCA) and statistical features. After that, convolutional neural networks (CNNs) became popular due to their locality preservation and translation invariance.
A 1D CNN is used to model sensor modality~\cite{zeng2014convolutional} while a 2D CNN regards the set of signals as an image and handles multichannel sensor readings~\cite{yang2015deep}. In order to capture the temporal dependencies of the sensor data, deep recurrent networks, especially long short-term memory cells (LSTMs), have achieved promising performance in HAR~\cite{ordonez2016deep, guan2017ensembles}. However, due to the complexity of LSTM, they require much labeled data to avoid overfitting. 

\subsection{Semi-Supervised Learning}
In semi-supervised learning, the model is trained on both labeled and unlabeled data \cite{chapelle2010semi}. 
Utilizing unlabeled data may improve a model's generalization ability.

Semi-supervised learning has been applied to HAR. An on-line adaptation method is proposed for semi-supervised learning for HAR \cite{cvetkovic2011semi}. The self-learning based approaches~\cite{stikic2008exploring, lopes2012semi} iteratively annotate the unlabeled data and selectively add them to the training dataset. The graph-based approach~\cite{stikic2009multi} connects labeled and unlabeled data and builds multiple graphs to propagate the labels based on similarity between features. However, these approaches treat the label propagation and classification as two separate processes. Thus, correlations between labeled data and unlabeled data may be ignored in the model. 



%

A recent semi-supervised method, ladder networks~\cite{rasmus2015semi}, can simultaneously train a deep auto encoder on an unlabeled dataset and a neural network on a labeled dataset. The ladder network shows superior performance in semi-supervised image classification for the MNIST and CIFAR-10 dataset.

\begin{figure*}[t]
   \centering
\includegraphics[width=1\textwidth]{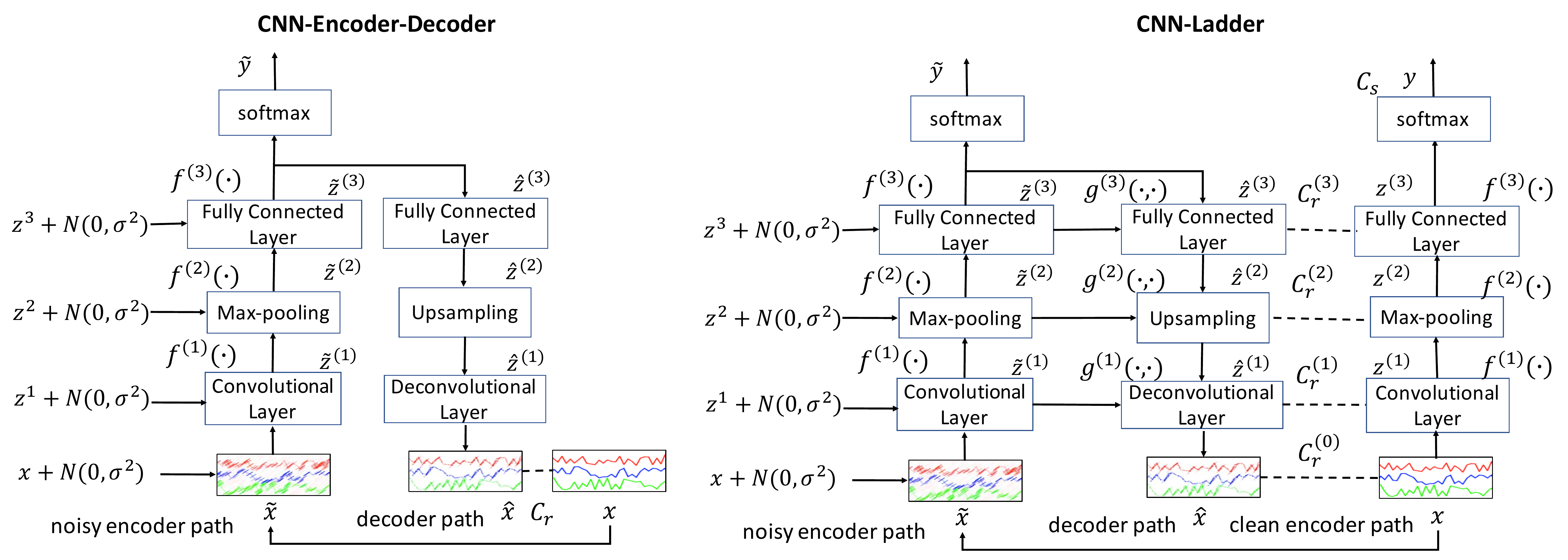} 
\caption{Structure of the CNN-Encoder-Decoder (left) and CNN-Ladder (right) applied to HAR. CNN-Ladder has two kinds of connections: \em{lateral connections} include $g^{(l)}(\cdot,\cdot)$ and reconstructed cost function $C_r^{(l)}$. \em{Vertical connections} contain clean encoder path ($x \rightarrow z^{(i)} \rightarrow y$), noisy encoder ($\tilde{x} \rightarrow \tilde{z}^{(i)} \rightarrow \tilde{y}$) path and decoder path ($\hat{z}^{(3)} \rightarrow \hat{z}^{(i)} \rightarrow \hat{x}$). The noisy encoder and clean encoder share the same mapping function $f$. The function $g$ is the denoising function, which is for reconstructing the clean input from high-level representation, $\hat{z}^{(3)}$. When we only consider the vertical connections and the lateral cost in the bottom $C_r^{(0)}$, the CNN-Ladder reduces to the CNN-Encoder-Decoder model (left).}
\label{fig:conv-ladder}
\end{figure*}
\section{Semi-Supervised CNN based Models}
We adopt the CNN since it provides stable latent representations at each network level, which preserved locality. It also has great potential to identify the various salient patterns of activity signals~\cite{yang2015deep}. We use the multi-sensor based CNN structure~\cite{yang2015deep} for both our supervised and semi-supervised learning approaches.

\subsection{CNN for Supervised Learning}
\label{sec:cnnsuper}
Consider a dataset with $N$ {\em labeled} sliding windows $(x_1, t_1)$,$(x_2,t_2)$,$...$,$(x_N,t_N)$, where $x_i$ is a sliding window input with length $T$ and $t_i$ is the activity label. A CNN maps the input $x_i=[x_{i1}, x_{i2},..x_{iT}]$ to hidden values $z_i^l=[z^l_{i1}, z^l_{i2},...,z^l_{id}]$ by convolutional kernels (to be learned in the training phase), where $l$ denotes the $l$-th layer (the input $x_i$ is also regarded as $0$-th layer, $z^0$ ). The CNN structure can be represented as:

\begin{align}
z_i^{(1)},...,z_i^{(L)}, y_i = \textnormal{CNN}(x_i),
\end{align}
where $\textnormal{CNN}(\cdot)$ contains at least one temporal convolutional layer, one pooling layer, and at least one fully connected layer prior a top-level softmax classifier. Then the supervised CNN cost function is of the form:
\begin{equation}
C_s = -\frac{1}{N} \sum_{i=1}^{N}\log{P(y_i = t_i | x_i)}.
\end{equation}
It requires a lot of labeled data to train a good CNN model.

\subsection{CNN Encoder-Decoder for Unsupervised Learning}
Assume that we also have $M$ {\em unlabeled} examples $x_{N+1}, x_{N+2},...,x_{N+M}$. The CNN-Encoder-Decoder consists of an encoder mapping $f$ and a decoder mapping $g$. The encoder adopts the CNN feed-forward process while the decoder contains upsampling and convolution operations. Our encoder-decoder structure is similar to a denoising autoencoder (DAE)~\cite{vincent2010stacked}. In the training, noise is injected into each layer in the network (including the input layer). The CNN-Encoder-Decoder minimizes the difference between the clean input $x_i$ and the reconstructed decoder output $\hat{x}_i$. Therefore, we have the cost function:
\begin{align}
C_r^{(0)} = \frac{\lambda}{M}\sum_{i=N+1}^{N+M} ||\hat{x}_i-x_i||_2^2,
\end{align}
where $\hat{x}_i$ is the reconstructed input. The decoder in the CNN-Encoder-Decoder~\cite{badrinarayanan2015segnet} contains upsampling for max-pooling decoding and another convolutional operation for deconvolution. The upsampling uses the memorized max-pooling indices from the corresponding encoder feature map(s) to produce sparse feature maps(s) as an input of the convolutional layer in the CNN-Encoder-Decoder~\cite{badrinarayanan2015segnet}. Then the sparse features are convolved with a trainable decoder filter bank to produce dense features.

\subsection{Semi-Supervised CNN-Encoder-Decoder for HAR}
We combine the supervised CNN and CNN-Encoder-Decoder to perform semi-supervised learning for HAR. Besides a set of labeled pairs $\{(x_i, t_i)\ | 1 \le i \le N \}$, semi-supervised learning~\cite{chapelle2010semi} uses unlabeled data $\{x_{i} | N+1 \le i \le N+M \}$ to help in training a classifier. 
 
In the case of a semi-supervised CNN-Encoder-Decoder, there are three paths for the labeled and unlabeled data: The clean encoding, noisy encoding, and the decoding:
\begin{align}
& z^{(1)}_i,...,z^{(L)}_i = \textnormal{Encoder}_{clean}(x_i) \label{eq_clean_encoder} \\
& \tilde{x}_i, \tilde{z}^{(1)}_i,...,\tilde{z}^{(L)}_i = \textnormal{Encoder}_{noisy}(x_i)  \label{eq_noisy_encoder} \\
& \hat{x}_i = \textnormal{Decoder}(\tilde{z}^{(L)}_i). \label{eq_decoder}
\end{align}

Both labeled and unlabeled clean data pass through the clean encoder path to compute hidden variables in the middle layers, $z_i^l$. For the noisy encoder path, both labeled and unlabeled data are corrupted by Gaussian noise and then transformed to a more abstract representation, $\tilde{z}_i^l$, by the noisy encoder. For labeled data ($\tilde{x}_i, 1 \le i \le N$), we carry out the prediction for labeled data on the top-level softmax classifier based on cross entropy cost. The predicted label is denoted by $\tilde{y}_i$. For the noisy unlabeled data ($\tilde{x}_i, N+1 \le i \le N+M$), the decoder tries to reconstruct it ($\hat{x}_i$) to be the same as the corresponding clean input ($x_i$). We use square error to evaluate this reconstruction error. The clean and noisy encoder paths share the same parameters, only the inputs are different in Fig~\ref{fig:conv-ladder}. (When we only consider the vertical connections and the lateral cost, CNN-Ladder in Fig~\ref{fig:conv-ladder} reduces to CNN-Encoder-Decoder.)

The CNN-Encoder-Decoder the cost function involves the supervised cross entropy cost from labeled data in the supervised CNN and the unsupervised denoising square error cost between the clean input and its noisy reconstruction output. Thus the cost function is 
\begin{align}
& C_e=  C_{s} + \lambda C_{r}^{(0)} \nonumber \\
& = - \frac{1}{N}\sum_{i=1}^{N}\log{P(\tilde{y}_i = t_i | x_i)} + \frac{\lambda}{M}\sum_{i=N+1}^{N+M} ||\hat{x}_i-x_i||_2^2,
\end{align}
where the supervised cost $C_{s}$ is the averaged cross entropy of the noisy output $\tilde{y}_i$ matching the target $t_i$ given the input $x_i$. The unsupervised cost $C_r$ is the averaged square error between the reconstruction output $\hat{x}_i$ and the clean input $x_i$. By using a semi-supervised CNN-Encoder-Decoder, we can potentially learn the network and features simultaneously from the data.

\subsection{Semi-Supervised CNN-Ladder for HAR} 
The semi-supervised Convolutional Ladder Network (CNN-Ladder) contains two kinds of connections: the vertical connections and the lateral connections (Fig~\ref{fig:conv-ladder}). The vertical connections have clean and noisy encoders (Eq~\ref{eq_clean_encoder}, Eq~\ref{eq_noisy_encoder}) and a decoder. The reconstruction $\hat{z}_i^{(l)}$ in the decoder is not only inferred from the upper layer $\hat{z}_i^{(l+1)}$, but also estimated from its corresponding layer in the noisy encoder. The estimation is a linear function $\hat{z}^{(l)} = g(\tilde{z}^{(l)}, \hat{z}^{(l+1)})$, where $\tilde{z}^{(l)}$ is the lateral noisy signal in the encoder and $\hat{z}^{(l+1)}$ is the reconstruction of its upper layer by batch normalization~\cite{rasmus2015semi}. These vertical skip-connections enable us to find better middle-level representations compared to regular encoder-decoder structures.

To improve the middle-level features reconstruction in the CNN-Encoder-Decoder, we also force the intermediate layers in the decoder to be similar to the corresponding layers in the encoder. In other words, the cost function of CNN-Ladder is
\begin{align}
\label{eq:ladder}
C_{l}& = C_s  + \sum_{l=0}^L \lambda_l C_{r}^{(l)} = - \frac{1}{N}\sum_{i=1}^{N}\log{P(\tilde{y}_i = t_i | x_i)} \nonumber \\
&\quad + \frac{1}{M}\sum_{i=N+1}^{N+M}\sum_{l=0}^L \lambda_l ||\hat{z}_i^{(l)}-z_i^{(l)}||_2^2 
\end{align}

If we train neural networks on limited unlabeled data, learned hidden features may have high variance and can be unstable. With the constraints from the lateral connection, the CNN-Ladder makes every layer, $C_r^{(l)}$, contribute to the cost function. As a result, more stable hidden features can be learned from large amount of unlabeled data. 
Stable hidden features can generate accurate representation of the middle level features, and lead to precise recognition of complicated activities.
For example, jumping jack activity prediction relies on stable and accurate representation of sub-components (spreading hands and legs, and clapping hands).

\section{Experiments}
\label{sec:experiments}
We validate our HAR approaches on three public datasets. First, we compare our methods to other neural network methods for HAR in a supervised learning setting. Second, we compare our methods to traditional semi-supervised learning methods for HAR. Third, we conducted experiments with varying amounts of labeled and unlabeled data, to understand the usability of our methods. Fourth, we discuss why our methods perform better than traditional semi-supervised learning methods in utilizing the unlabeled data for semi-supervised HAR. Deep learning (CNN, Pretrained CNN, Pseudo-label CNN, CNN-Encoder-Decoder, CNN-Ladder) is performed on a server equipped with a Tesla K20c GPU and 64G memory. The traditional learning algorithms (LR, Self-training) are run on the same server with an Intel Xeon E5 CPU. The implementation of CNN-Ladder is based on the Ladder Networks.\footnote{https://github.com/CuriousAI/ladder}

\subsection{Datasets}
The raw sensor data is segmented by a common sliding window technique. The window size is $2$ seconds with $50\%$ overlap. Data within each window is denoted as an \emph{example}. All the results are averaged using leave-one-subject-out cross validation. To ensure that labeled training data includes all the activity classes, we construct balanced labeled training datasets. The datasets used are as follows.

The \textbf{ActiTracker}~\cite{kwapisz2011activity} dataset contains 6 daily activities collected in a controlled laboratory environment. The activities are ``jogging,'' ``walking,'' ``ascending stairs,'' ``descending stairs,'' ``sitting'' and ``standing.'' The data are recorded from 36 users, with a 20Hz sampling rate resulting in 1,098,207 examples. After segmentation, there are around 110,000 examples (sliding windows). The number of examples for testing varies from 1,000 to 5,000.

The \textbf{PAMAP2} dataset ~\cite{reiss2012introducing} consists of 12 lifestyle activities (``walking,'' ``lying down,'' ``knees bending,'' etc.) by 9 participants. Accelerometer, gyroscope, magnetometer, temperature, and heart rate data are recorded from inertial measurement units located on the hand, chest and ankle over 10 hours, resulting in 52 dimensions. The number of examples is 
3,850,505. To have a temporal resolution comparable to the ActiTracker dataset, we downsampled the data to 33.3Hz, resulting in around 33,000 examples. The number of examples of test data in each experiment is around 4,500.


The \textbf{mHealth} dataset ~\cite{banos2014mhealthdroid} contains recordings from 10 participants while performing 12 physical activities, including daily life activities (``standing,'' ``lying down,'' etc.) and exercise activities (``cycling,'' ``jogging,'' etc). Accelerometers, gyroscopes, magnetometer and ECG data are recorded from inertial measurement units placed on a participant's chest, right wrist and left ankle. The data has 43,744 examples with 23 dimensions. In our experiment, we downsampled the data to 20Hz, resulting in around 8,000 examples. The number of examples used for testing is around 1,000.

\subsection{Experimental Setup}
We consider these supervised learning baselines:
\begin{itemize}
\item \textbf{Logistic Regression (LR)}~\cite{bao2004activity}: We using traditional logistic regression for supervised learning in combination with statistical features (mean, standard deviation, correlation, max, min). 
\item \textbf{Supervised Convolutional neural network (CNN)}~\cite{yang2015deep}: The structure of the supervised CNN is the same as the clean path in our CNN-Ladder.\footnotemark[3]

\footnotetext[3]{Network structure: convv:40:5:1:1-maxpool:2:2-convv:50:3:1:1-maxpool:2:2-convv:20:3:1:1-convv:50:1:1:1-fc.~\cite{rasmus2015semi}}
\end{itemize}
We also study traditional semi-supervised learning baselines.

\begin{itemize}
\item \textbf{Unsupervised Pretrained CNN}~\cite{erhan2009difficulty}: The pretrained CNN uses the unlabeled data to initialize the network parameters. We use an
unsupervised pretraining method similar to multi-layer perceptron (MLP) pretraining. In the first step, the pretrained CNN uses the unlabeled data to perform encoding and decoding with the CNN structure to initialize the parameters of the network. 

\item \textbf{Self-training method with logistic regression (Self-training)}~\cite{stikic2008exploring}: In self-training, an LR classifier is first trained using (a small amount of) labeled data. Then we use the trained LR model to predict the labels of unlabeled data. In each iteration, predictions with high confidence are added to the labeled training set, where these predictions are now considered as the labels. In our experiments, the confidence threshold is $0.95$.

\item \textbf{Pseudo-label}~\cite{lee2013pseudo}: The pseduo-label approach is essentially a self-training method. The predicted labels of the unlabeled data are used in a fine-tuning phase to improve the recognition performance.
\end{itemize}



A result is averaged across all leave-one-subject-out cross validation experiments. Thus, in each experiment, we use one user for test and the rest of the users for training. We evaluate the results using mean F1-score because the activity datasets are highly biased. The F1-score is a harmonic mean of precision and recall. The \textbf{mean F1-score}, $F_m$, is the mean F1-score across all the classes:
\begin{equation}
F_m = \frac{2 \cdot \text{precision} \cdot \text{recall} }{\text{precision} + \text{recall}}
\end{equation}
where for a given class
\begin{equation*}
\text{precision} = \frac{TP}{TP+FP}, \qquad \text{recall} = \frac{TP}{TP+FN}.
\end{equation*}
Here, $FP$ and $FN$ are counts of False
Positives and False Negatives, respectively.

Table \ref{reproduce} shows that our baseline of supervised CNN is comparable to the results in previous papers. We also evaluate our CNN baseline on all users, instead of using the setting in the previous works. The mean F1 scores are $79.54$, $75.38$ and $92.83$ on ActiTracker, PAMAP2 and mHealth, respectively.

\begin{table}[t]
\centering
\label{my-label}
\begin{tabular}{|c|c|c|c|c|}
\hline
            & \multicolumn{2}{c|}{\begin{tabular}[c]{@{}c@{}}Previous\\ Papers\footnotemark[1] \footnotemark[2] \footnotemark[3] \end{tabular}} & \multicolumn{2}{c|}{\begin{tabular}[c]{@{}c@{}}This\\ Paper\end{tabular}}  \\ \hline
Data        & LR               & CNN              & LR                                          & CNN                                                 \\ \hline
ActiTracker (Accuracy) & 78.10\footnotemark[1]             & 90.88\footnotemark[2]            & 89.27                                       & 93.84                                              \\ \hline
PAMAP2 ($F_m$)      & -                & 93.70\footnotemark[3]            & 86.86                                       & 92.24                                                \\ \hline
\end{tabular}
\caption{We reproduce the results of LR and CNN on ActiTracker and PAMAP2, reported in [Kwapisz et al. 2011, Zeng et al. 2014, Hammerla et al. 2016]. We show the results under the previous papers' settings. On ActiTracker, the results are in accuracy. On PAMAP2, the results are in $F_m$. 
\label{reproduce}}
\end{table}


\footnotetext[1]{We only carry out 10-fold cross validation for [Kwapisz et al. 2011].}
\footnotetext[2]{We only carry out 10-fold cross validation for [Zeng et al. 2014].}
\footnotetext[3]{User 6 is for the test set, user 5 is for the validation set and the rest of the users are used for the training set [Hammerla et al. 2016].}

\subsection{Comparing with Supervised Methods}
\label{sec:accimprove}
We compare our methods to several supervised methods, to study how our methods utilize unlabeled data in HAR.
The baseline methods LR and CNNs do not use unlabeled data.

The results are shown in Table~\ref{tab:main}. On all the three datasets, CNN-Encoder-Decoder and CNN-Ladder perform consistently better than LR and CNN. In particular, CNN-Ladder achieves $17.64\%$, $3.59\%$, $9.65\%$ improvements in mean F1-score on the three datasets, compared to the best of LR and CNN. Second, CNN-Ladder has higher $F_m$ score than CNN-Encoder-Decoder on the three datasets.



\begin{table*}[]
\centering
\label{tab:main}
\begin{tabular}{|l||r|r||r|r|r||r|r||r|r|}
\hline
            & \multicolumn{2}{c||}{Supervised} & \multicolumn{3}{c||}{Semi-Supervised}          & \multicolumn{2}{c||}{Our Semi-Supervised} & \multicolumn{2}{c|}{Improvement}             \\ \hline
            & LR             & CNN            & \makecell{Pretrained \\CNN} & Self-Training & Pseudo-Label & \makecell{CNN-Encoder\\-Decoder}         & CNN-Ladder         & $\Delta$Supervised & \makecell{$\Delta$Semi-\\Supervised} \\ \hline
ActiTracker & 39.34          & \circletext{48.68}          & \circletext{49.86}          & 41.52         & 46.00        & 63.58                       & \circletext{66.32}              & 17.64              & 16.46                   \\ \hline
PAMAP2      & \circletext{51.31}          & 50.22          & 48.54          & 47.86         & \circletext{50.79}        & 52.68                       & \circletext{54.90}              & 3.59               & 4.11                    \\ \hline
mHealth     & 57.73          & \circletext{59.73}          & \circletext{60.88}          & 59.43         & 60.31        & 66.61                       & \circletext{69.38}              & 9.65               & 8.50                    \\ \hline
\end{tabular}
\caption{The $F_m$ score of supervised methods (LR and CNN), traditional semi-supervised methods (Self-training and Pseudo-label) and our presented methods (CNN-Encoder-Decoder and CNN-Ladder). We circle the best $F_m$ scores from supervised, semi-supervised and our semi-supervised approaches, respectively. Both of our methods (CNN-Encoder-Decoder, CNN-Ladder) are significantly better compared to the CNN and the Pretrained CNN with $p$-value $< 0.05$. \label{tab:main}}
\end{table*}

Those results suggest that CNN-Encoder-Decoder and CNN-Ladder can effectively make use of the unlabeled data, to significantly improve accuracy. CNN-Ladder performs better than CNN-Encoder-Decoder, perhaps because better hidden features are trained. In CNN-Ladder, the loss function considers the difference between each layer of CNN and its decoder, while CNN-Encoder-Decoder only considers the difference between the final reconstructed output and the original input.


\subsection{Comparing with Traditional Semi-supervised Methods}
We compare our methods to traditional semi-supervised methods, to study how our methods can utilize the same unlabeled data to achieve more accurate predictions.


The comparisons between Pretrained CNN, Self-Training, Pseudo-Label, CNN-Encoder-Decoder, and CNN-Ladder are shown in Table \ref{tab:main}. It can be observed that CNN-Encoder-Decoder and CNN-Ladder perform better than Pretrained CNN, Self-Training, and Pseudo-Label. Specifically, CNN-Ladder achieves about $16.46\%$, $4.11\%$, $8.5\%$ improvements in mean F1-scores on the three datasets, compared to the best of Pretrained CNN, Self-Training, and Pseudo-Label. 


One disadvantage of Self-Training and Pseudo-Label that we observed is that these iterative methods need careful selection of the confidence threshold. If the confidence threshold is not appropriately selected, some unlabeled data will be assigned wrong labels and the errors will propagate in later iterations.
However, in semi-supervised CNNs, no confidence threshold is needed and all available unlabeled data are input together with labeled data to train the models. Without using confidence thresholds, training neural network requires less domain knowledge and is much easier compared to training Self-Training and Pseudo-Label models.

\begin{figure*}[ht]
   \centering
        \includegraphics[width=0.31\textwidth]{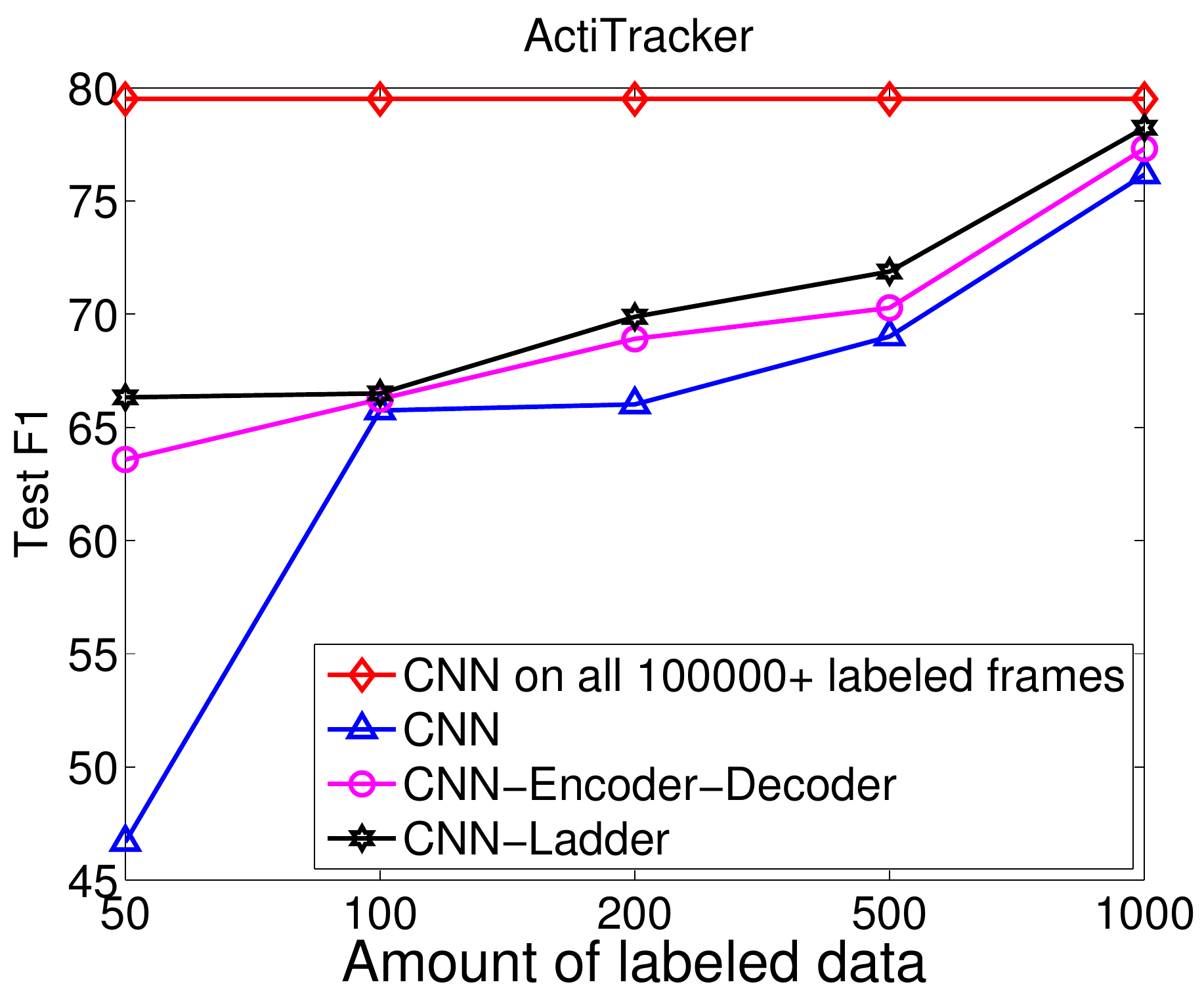} 
         \includegraphics[width=0.31\textwidth]{trend1-eps-converted-to.pdf} 
       \includegraphics[width=0.31\textwidth]{trend1-eps-converted-to.pdf} 
        \caption{The $F_m$ scores of CNN, CNN-Encoder-Decoder, and CNN-Ladder, with varying number of labeled examples. 
        The $F_m$ scores of supervised CNN on all labeled training examples are also shown as red lines.}\label{fig:label}
\end{figure*}

\subsection{Varying Amount of Labeled Data} 
In this section, we study the performance of our models trained with varying amounts of {\em labeled} data. We evaluate the $F_m$ score of supervised CNN, CNN-Encoder-Decoder and CNN-Ladder trained on $50$, $100$, $200$, $500$, and $1$,$000$ labeled examples. The rest of the samples in the training set are regarded as unlabeled.


Figure \ref{fig:label} shows the $F_m$ trend when we vary the number of labeled examples. There are three observations. First, the $F_m$ scores of supervised CNN, CNN-Encoder-Decoder and CNN-Ladder generally improve when we have more labeled examples. Second, with the same number of labeled examples, CNN-Encoder-Decoder, and CNN-Ladder usually achieves higher $F_m$ scores than CNN. Third, when CNN-Ladder is learned from $1$,$000$ examples, its mean $F_m$ score is already very competitive with supervised CNN learned from more than $100$,$000$, $10$,$000$, and $8$,$000$ labeled examples from ActiTracker, PAMAP2 and mHealth, respectively. These results indicate that compared to CNN, CNN-Ladder can achieve similar accuracy but with much smaller number of labeled examples.

\subsection{Varying Amount of Unlabeled Data} 
We now study the performance of our models trained with varying amounts of {\em unlabeled} data. We evaluate the $F_m$ score of supervised CNN, CNN-Encoder-Decoder, and CNN-Ladder trained on $50$ labeled examples and varying amounts of unlabeled examples. On ActiTracker, the number of unlabeled examples varies from $100$ to $50$,$000$. On PAMAP2 and mHealth, the number  varies from $100$ to $10$,$000$, as these two datasets are relatively small.

\begin{figure*}[h]
   \centering
        \includegraphics[width=0.31\textwidth]{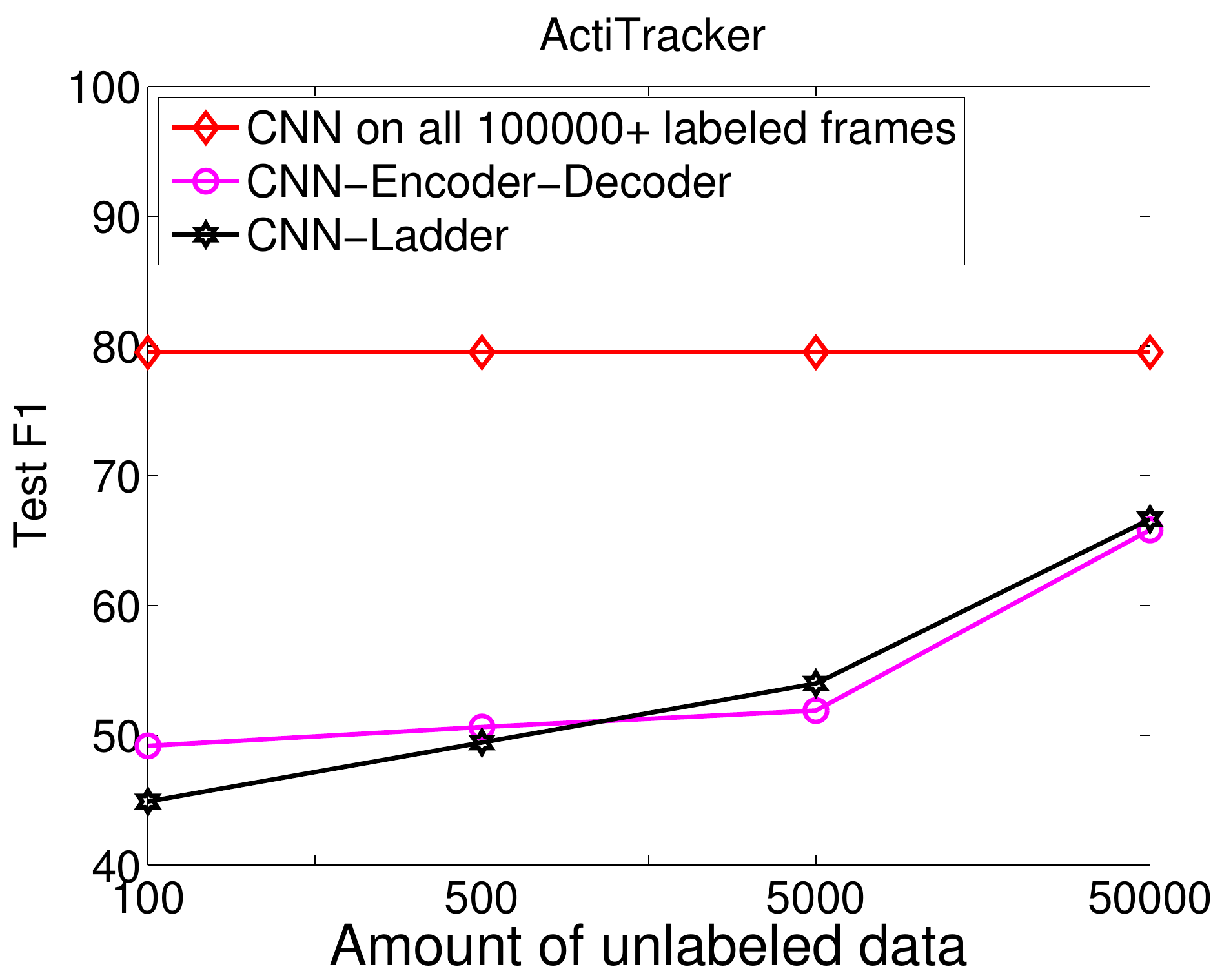}
        \includegraphics[width=0.31\textwidth]{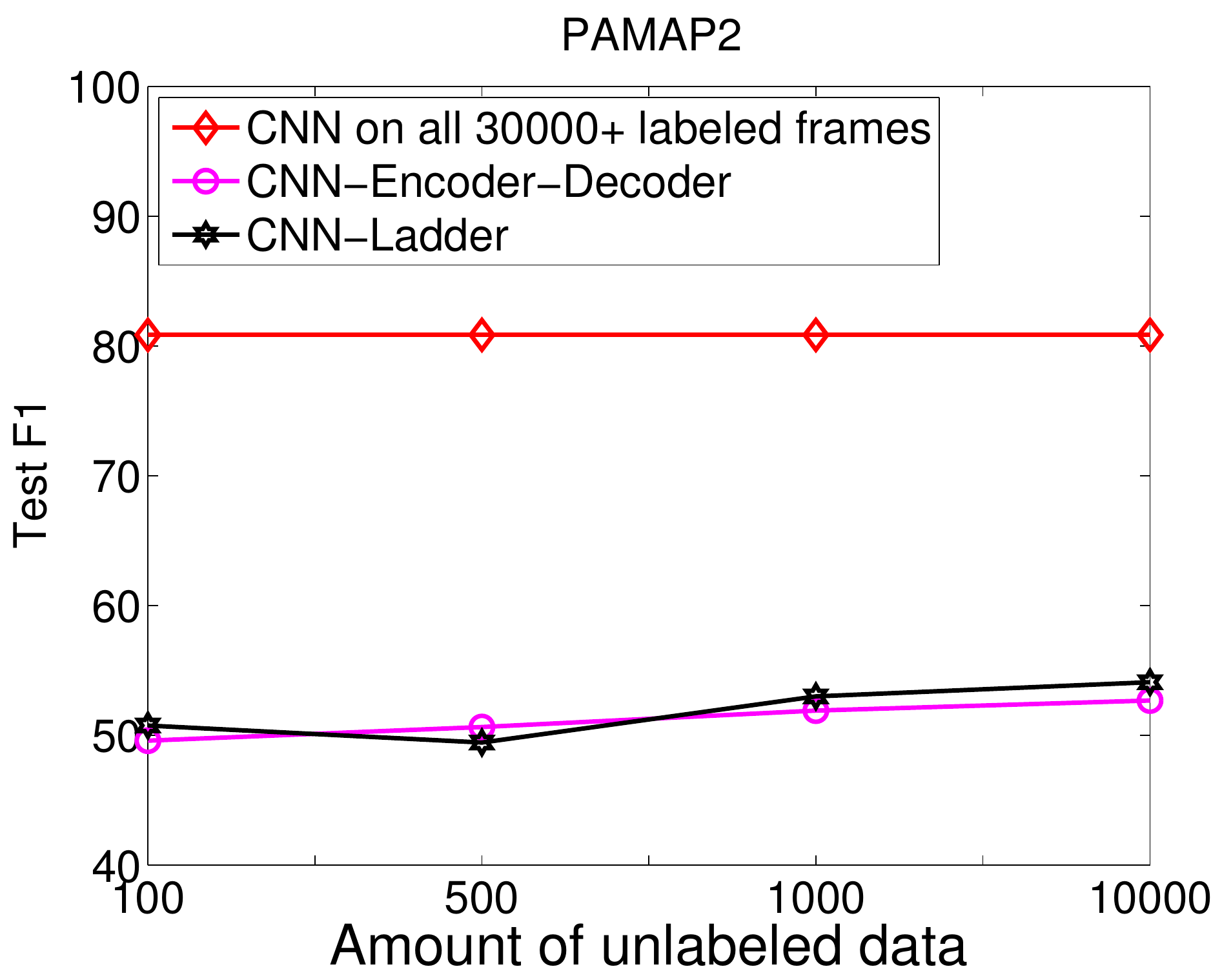}
        \includegraphics[width=0.31\textwidth]{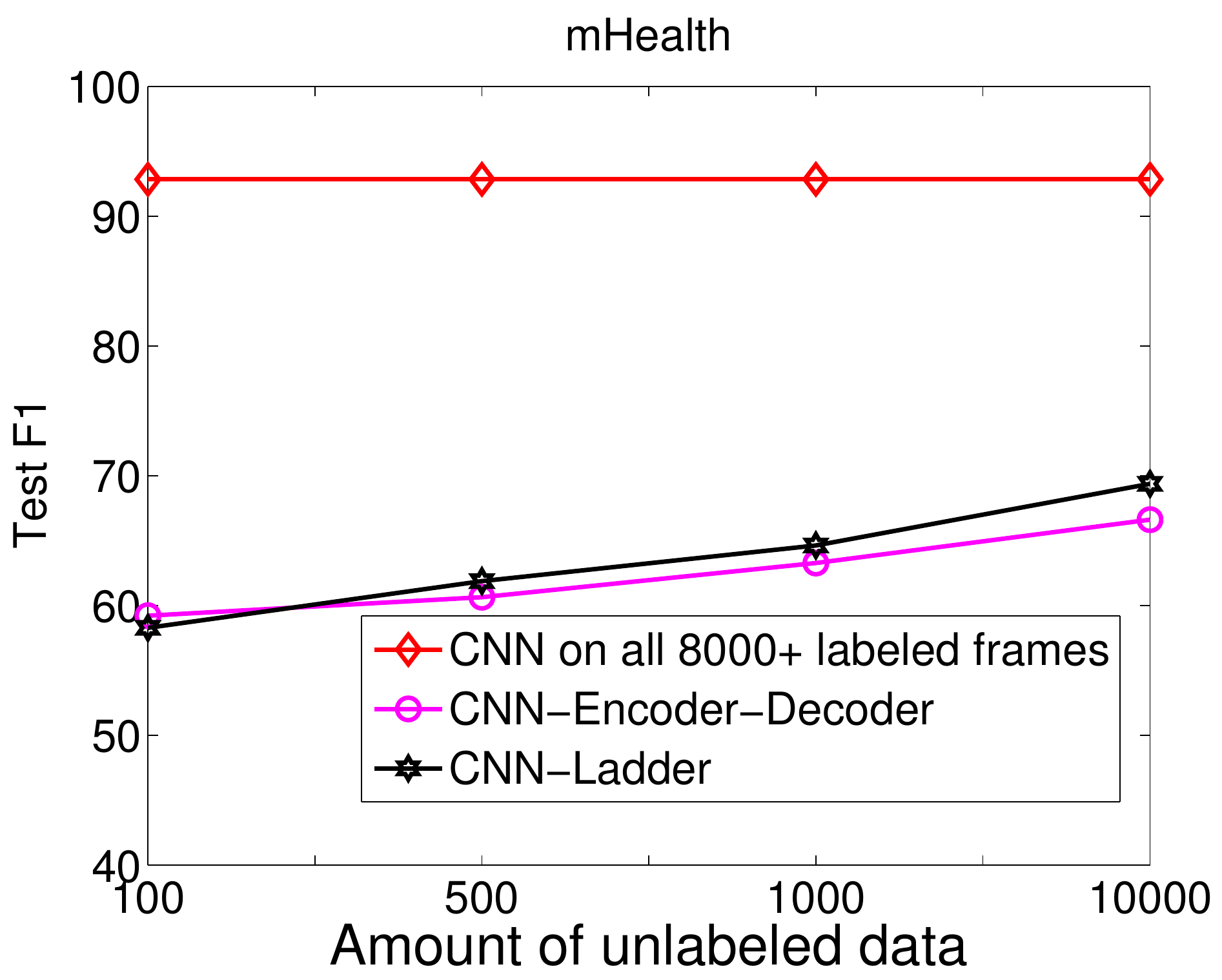}
        \caption{The $F_m$ scores of CNN-Ladder and CNN-Encoder-Decoder, with $50$ labeled examples and varying amount of unlabeled examples. The $F_m$ scores of supervised CNN on a large number of labeled examples are also shown as red lines. The result of CNN-Ladder is significantly better than the CNN approach with $p$-value $< 0.05$.
        }\label{fig:unlabel}
\end{figure*}

Figure \ref{fig:unlabel} shows the experimental results. With an increasing amount of unlabeled data, the $F_m$ score typically impoves for both CNN-Encoder-Decoder and CNN-Ladder. This suggests that better latent features in the auto-encoder can be trained with more unlabeled examples and help adjust the latent CNN features, thereby improving accuracy.

\subsection{The Impact of Adjusting Features in Different Layers}
\label{sec:impact}
We now study the importance of adjusting different layers in CNN-Ladder with unlabeled data. Specifically, we adjust the $\lambda_l$ of CNN-Ladder in Equation \ref{eq:ladder} to observe the impact of making the latent features between CNN and autoencoder more or less similar, in different layers. We run a set of experiments for different layers $l$, where $l \in \{0, 1, \cdots, L\}$. In each experiment, we emphasize layer $l$ by setting $\lambda_l = 1$ and $\lambda_k = 0.1$, where $k \in \{0, 1, \cdots, l-1\} \cup \{l , l+1, \cdots, L\}$. In our CNN-Ladder, $L =9$. 

\begin{figure}[h]
   \centering
        \includegraphics[width=0.33\textwidth]{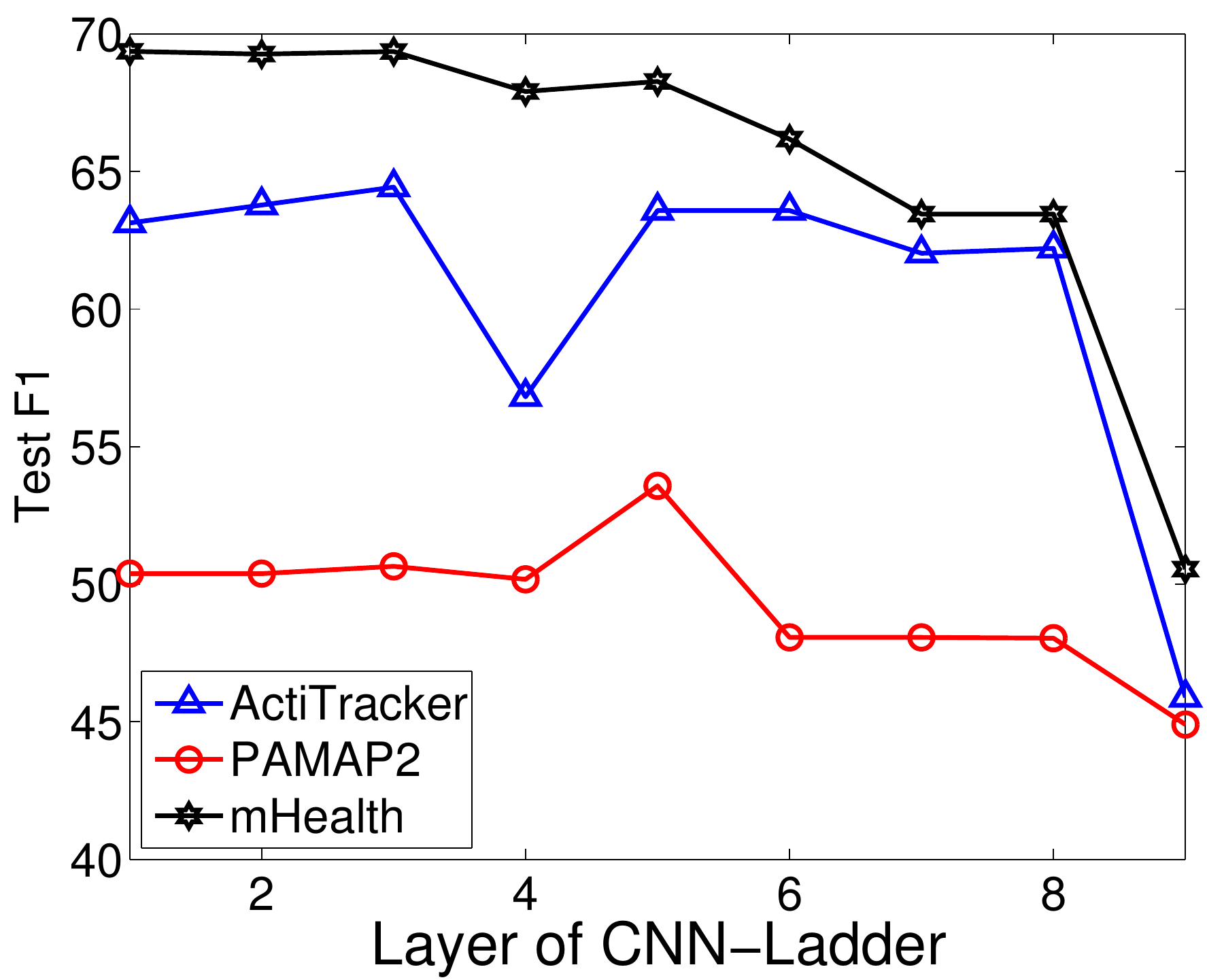} 
        \caption{The impact of making different layers' latent features of CNN and autoencoder very similar in CNN-Ladder. Utilizing the unlabeled data starting from very low-level features is very important in semi-supervised HAR, but beyond the scope of traditional semi-supervised learning methods.}\label{fig:layer}
\end{figure}

The resulting $F_m$ scores when varying the weights of different layers of CNN-Ladder are shown in Figure \ref{fig:layer}. 
A high $F_m$ score can typically be achieved by setting a large $\lambda_l$ for the layers representing low-level features. This indicates that low-level features of the neural networks can be much improved by using the unlabeled data. 


In contrast, utilizing the unlabeled data for low-level features is missed in traditional semi-supervised learning methods for HAR, such as Self-Training.
Self-Training uses the unlabeled data only after feature engineering is already done. That is, the handcrafted features are independent from whether unlabeled data is available or not.
In a similar way, low-level features of traditional neural network methods, such as CNN, may be not as good as the low-level features of CNN-Ladder in the semi-supervised HAR. 

\begin{figure*}[h]
   \centering
  \includegraphics[width=0.23\textwidth]{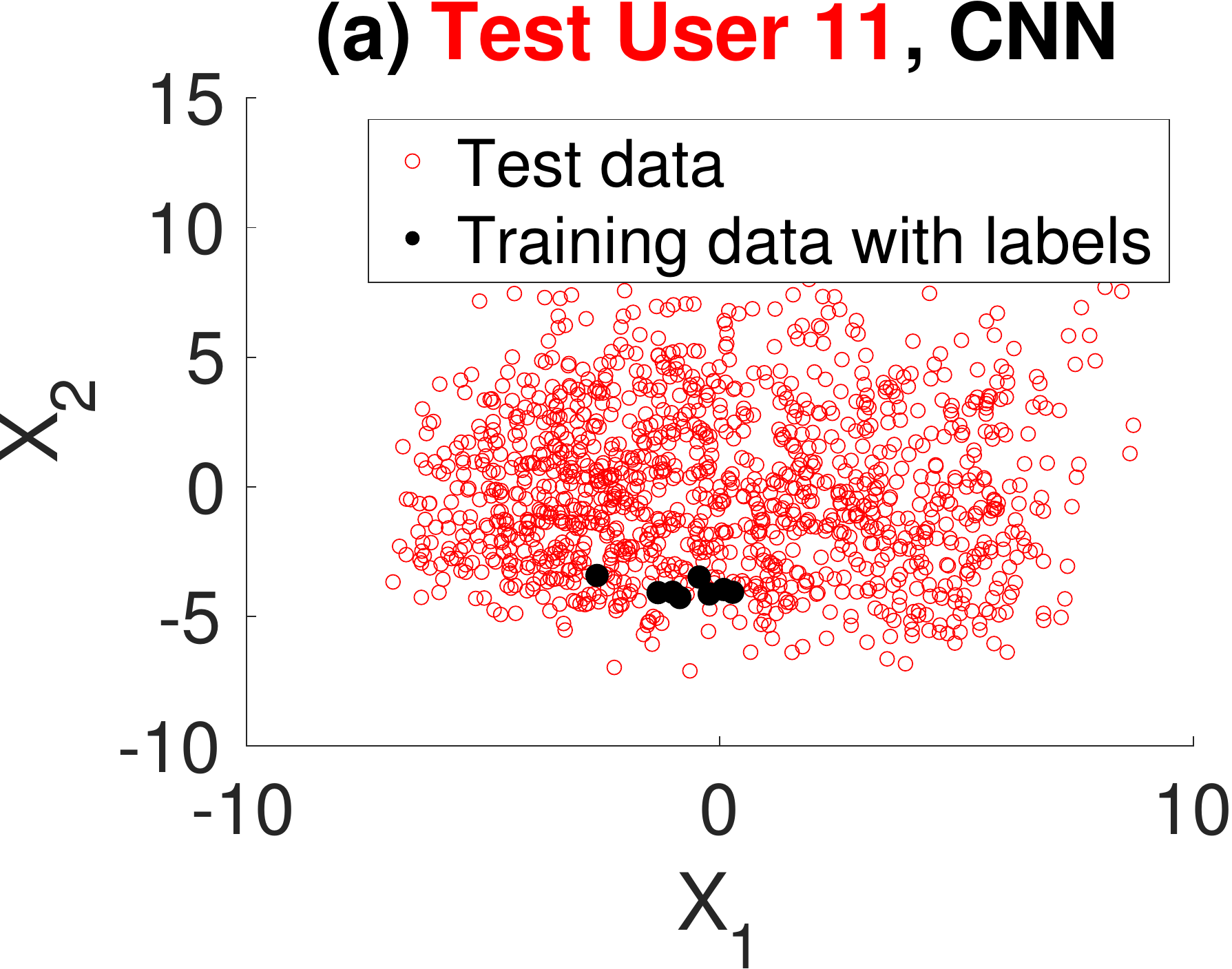} 
 \includegraphics[width=0.23\textwidth]{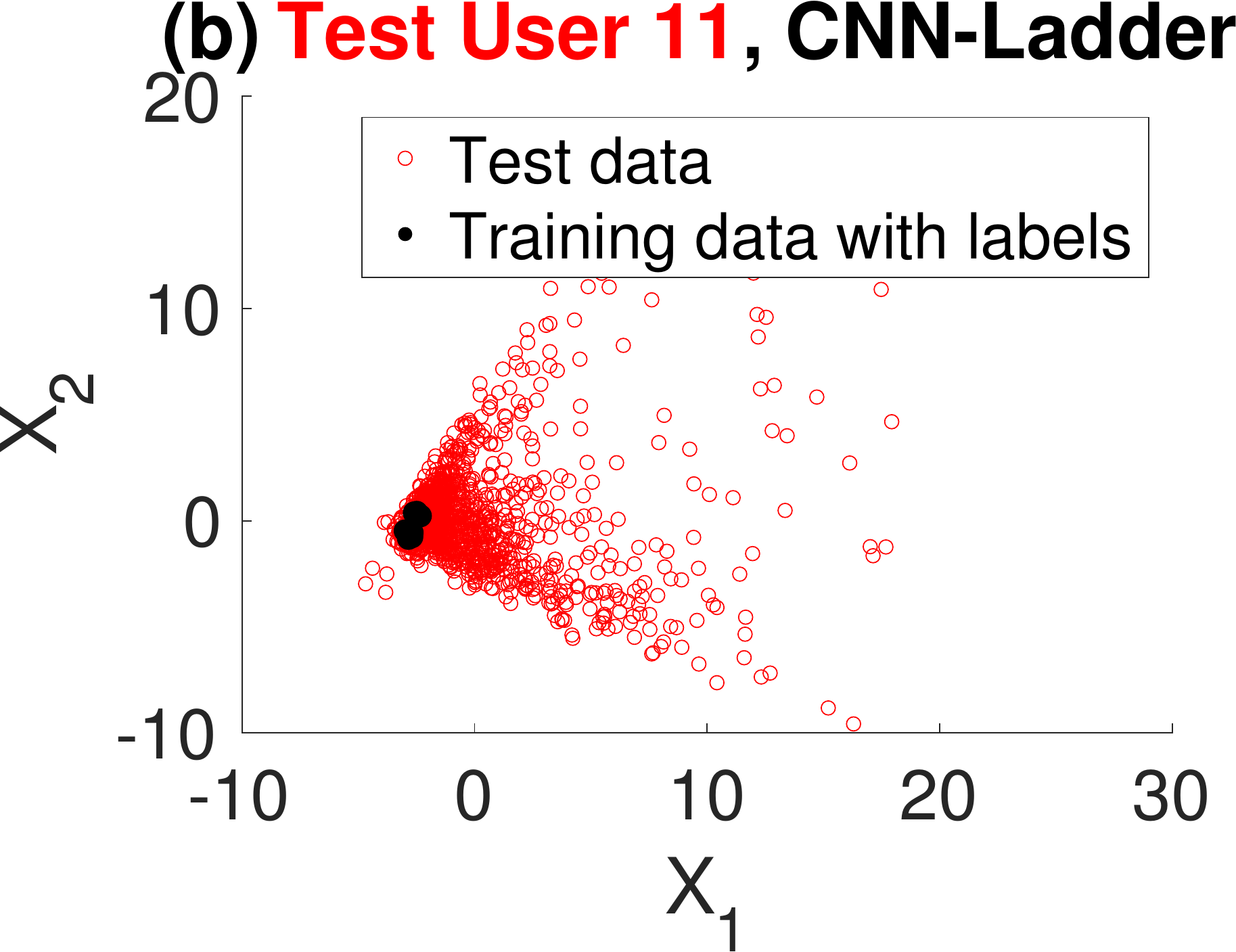} 
 \includegraphics[width=0.23\textwidth]{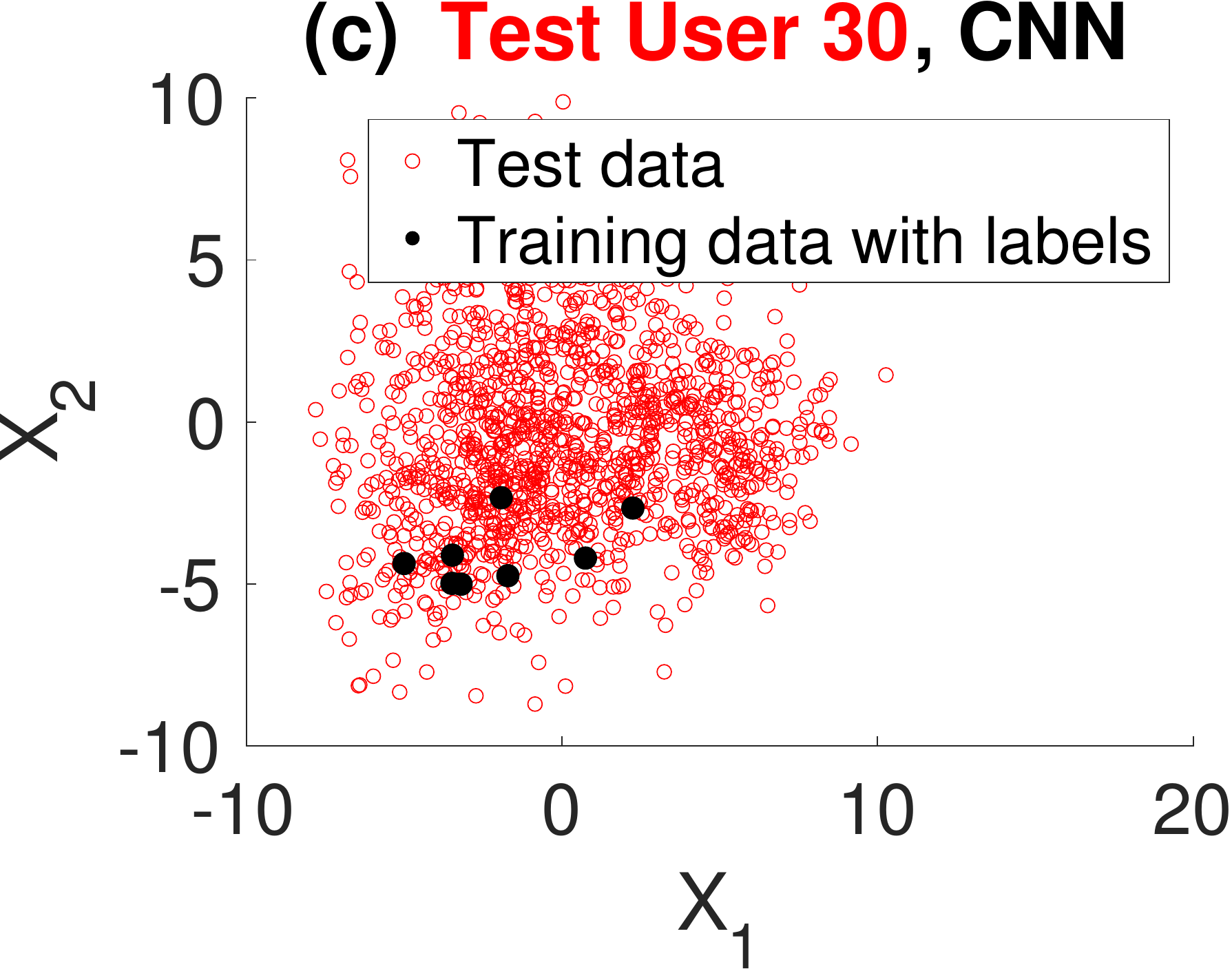} 
 \includegraphics[width=0.23\textwidth]{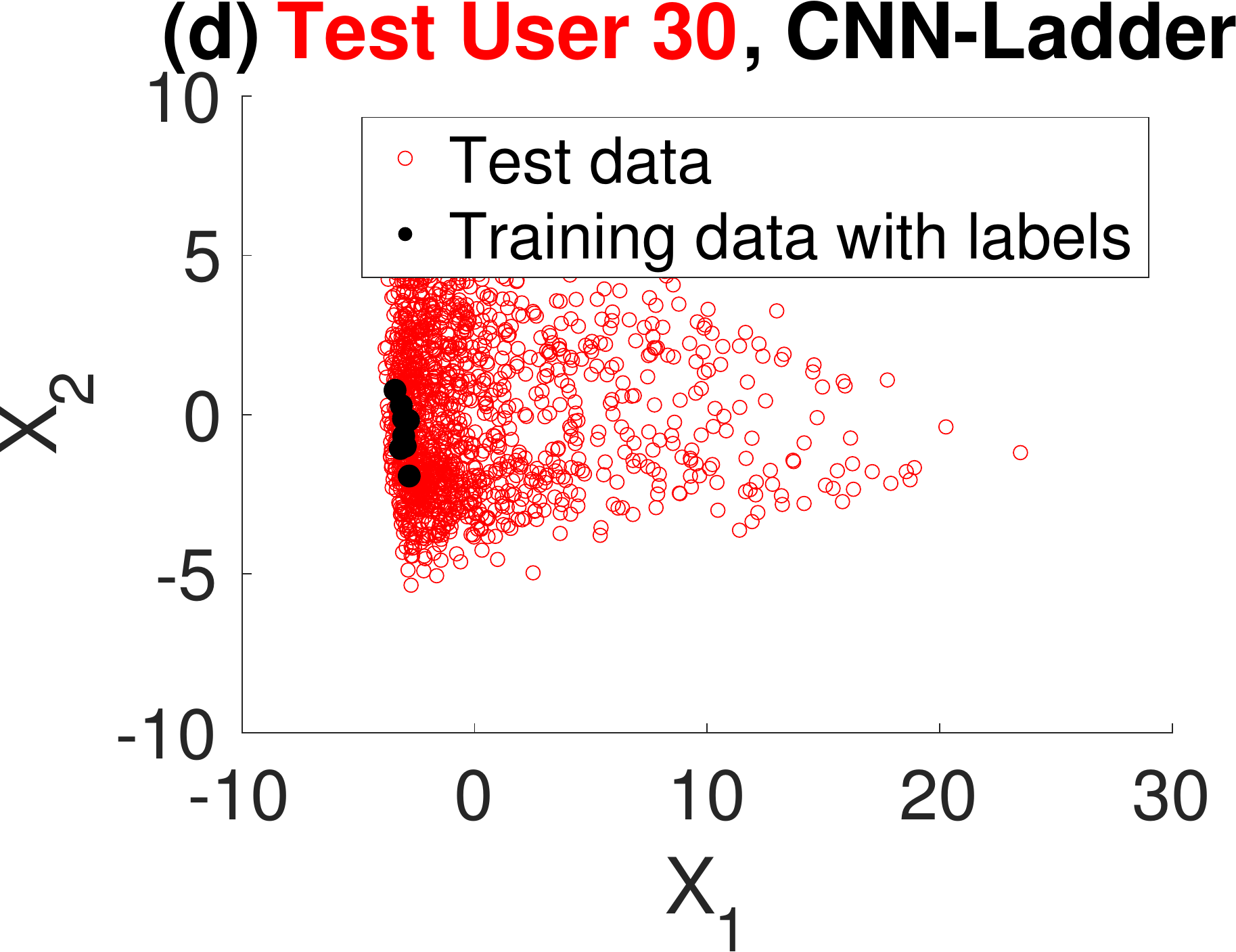} 
  \caption{The visualizations of the low-level features of traditional CNN and CNN-Ladder using PCA. The black dots are labeled data of jogging activity from users in the training set. The red dots are unlabeled data of the same activity from a different user not in the training set. Although the red and black dots belong to the same class, they are badly scattered in the traditional CNN. In CNN-Ladder, the red dots are more concentrated around the black dots.}\label{fig:vv}
\end{figure*}

\subsection{How Does CNN-Ladder Achieve Higher $F_m$?}
\label{sec_ladder_layer}
As discussed in Section \ref{sec:impact}, CNN-Ladder can adjust low-level features with unlabeled data, while traditional semi-supervised methods' low-level features are independent from unlabeled data.
This section seeks to better understand how CNN-Ladder's low-level features help achieve high $F_m$ scores in HAR.

We visualize the features in the last layer of (i) CNN-Ladder with unlabeled data versus (ii) CNN without unlabeled data. PCA is used to reduce the dimensionality of the data, and only the eigen-vectors with the largest two eigen-values are selected as axes in Figure \ref{fig:vv}. To understand how CNN-Ladder benefits from varying low-level features, we show two cases where CNN-Ladder achieves high $F_m$ score while CNN does not.  

In the prediction of jogging activity for User $11$, the features in the last layer of CNN-Ladder with unlabeled data and CNN without unlabeled data are shown in Figure \ref{fig:vv}(a) and \ref{fig:vv}(b). In this case, CNN fails to predict the jogging activity of different users as the same activity. This is caused by the varying behaviors of different users, especially when the labeled examples are limited as shown in Figure \ref{fig:vv}(a).
Interestingly, in the two-dimensional visualization of  features in CNN-Ladder in Figure \ref{fig:vv}(b), the test examples concentrate in the region where the labeled data locate. Using the low-level feature representations trained with additional unlabeled data, the jogging activities of different users become similar, even with differences in jogging behaviors between different users. Figure \ref{fig:vv}(c) and \ref{fig:vv}(d) show another similar case for User $30$.

The visualization results indicate that with unlabeled data, CNN-Ladder can learn discriminative high-level features even when labeled training data is very limited. Consequently, it is easier for CNN-Ladder to achieve higher $F_m$.

\section{Conclusion}
\label{sec:con}
We study the CNN-Encoder-Decoder and CNN-Ladder architectures for semi-supervised human activity recognition. 
The experimental results demonstrate that our proposed methods can achieve significant $F_m$ improvements, compared to supervised learning methods and traditional semi-supervised learning methods. We carefully study how CNN-Ladder achieves higher $F_m$ in human activity recognition. The empirical results show that it is very helpful to use unlabeled data to better learn low-level features in CNNs human activity recognition. 

\section{Acknowledgement}\label{sec:acknow}
This research is supported in part by the National Science Foundation under the award 1346066: ``SCH: INT: Collaborative Research: FITTLE+: Theory and Models for Smartphone Ecological Momentary Intervention''.





%

\bibliographystyle{IEEEtran}
\bibliography{ladder}

\end{document}